\documentclass{article}
\usepackage[utf8]{inputenc}

\usepackage{amsmath}
\usepackage{amsfonts}
\usepackage{bm}

\usepackage{graphicx}

\usepackage{arxiv}
\usepackage{booktabs}
\usepackage{caption}
\usepackage{lscape}
\usepackage{natbib}
\usepackage{placeins}
\usepackage{subcaption}
\usepackage{url}

\newcommand{\Sidak}{\v Sid\' ak}

\title{Post-Selection Confidence Bounds for\\Prediction Performance}
\author{Pascal~Rink\thanks{Correspondence to: Pascal~Rink, p.rink@uni-bremen.de}\\
  Institute for Statistics and \\
  Competence Center for\\
	Clinical Trials Bremen\\
  University of Bremen\\
  Bremen, Germany \\  
   \And
  Werner~Brannath \\
  Institute for Statistics and\\ 
  Competence Center for\\
  Clinical Trials Bremen\\
  University of Bremen\\
  Bremen, Germany \\}
\date{\today}

\begin{document}

\maketitle

\begin{abstract}%
In machine learning, the selection of a promising model from a potentially large number of competing models and the assessment of its generalization performance are critical tasks that need careful consideration. Typically, model selection and evaluation are strictly separated endeavors, splitting the sample at hand into a training, validation, and evaluation set, and only compute a single confidence interval for the prediction performance of the final selected model. We however propose an algorithm how to compute valid lower confidence bounds for multiple models that have been selected based on their prediction performances in the evaluation set by interpreting the selection problem as a simultaneous inference problem. We use bootstrap tilting and a maxT-type multiplicity correction. The approach is universally applicable for any combination of prediction models, any model selection strategy, and any prediction performance measure that accepts weights. We conducted various simulation experiments which show that our proposed approach yields lower confidence bounds that are at least comparably good as bounds from standard approaches, and that reliably reach the nominal coverage probability. In addition, especially when sample size is small, our proposed approach yields better performing prediction models than the default selection of only one model for evaluation does. 
\end{abstract}


\section{Introduction} \label{sec:introduction}

Many machine learning applications involve both model selection and the assessment of that model's prediction performance on future observations. This is particularly challenging when only little data is available to perform both tasks. By allocating a greater fraction of the data towards model selection the goodness assessment gets less reliable, and allocation of a greater fraction towards goodness assessment poses the risk of selecting a sub-par prediction model. In such situations it is desirable to have a procedure at hand that resolves this problem reliably.

Recent work by \cite{Westphal-2020} showed that it is beneficial in terms of final model performance and statistical power to select multiple models for goodness assessment in spite of the need to correct for multiplicity then. While \cite{Westphal-2020} proposed a multiple test in such cases, we here propose a way how to compute valid lower confidence bounds for the conditional prediction performance of the final selected model. Note that reporting a confidence interval here perfectly makes sense since a point estimate for the performance does not incorporate the uncertainty of estimation from an evaluation set at all. 

We follow the idea of \cite{Berk-2013} and interpret this post-selection inference problem as a simultaneous inference problem, controlling for the family-wise error rate 
\begin{equation} \label{eq:fwer}
P_\theta ( \theta_i < \theta_{i, L} \text{ for any } i \in \{1, \ldots, m\} ) \leq \alpha, 
\end{equation}
where $\theta_i$ denotes the performance of prediction model $i$, $\theta_{i, L}$ denotes the corresponding lower confidence bound at significance level $\alpha > 0$, and $\theta$ is the true predictive performance. With this type 1 error control, in practice, we are therefore able to answer the question whether there is a model among the competition that has prediction performance $\theta_i$ at least as large as a reference performance $\theta_0$ with high confidence, no matter how and which subset of the initial competition has been selected for evaluation. In particular, since 
\begin{equation} \label{eq:from-fwer-follows-marginal}
\alpha \geq P_\theta (\cup_{i = 1}^m \{ \theta_i < \theta_{i, L} \}) \geq P_\theta (\theta_s < \theta_{s, L}), 
\end{equation}
for some $s \in \{1, \ldots, m\}$, this coverage guarantee carries over to a potentially final selected model $s$. 

This might be an overly eager requirement to meet in certain cases and might lead to somewhat conservative decisions. Therefore, in order to increase power, we do not evaluate all of the candidate models, but only a promising selection of them. Or to put it the other way around: we exclude models from evaluation that are likely not going to be the best performing ones. However, in principle, it is also possible to report lower confidence bounds for the prediction performance of all the competing models that remain universally valid in the sense that they work with any measure of prediction performance (as long as it accepts weights), with any combination of prediction models even from different model classes, any model selection strategy, and are computationally undemanding as no additional model training is involved. While \cite{Berk-2013} proposed universally valid post-selection confidence bounds for regression coefficients, we are interested in a post-selection lower confidence bound for the conditional prediction performance of a model selected based on its evaluation performance. 

\subsection{Conditional vs Unconditional Performance}

\begin{figure}
\begin{center}
\includegraphics[width=\textwidth]{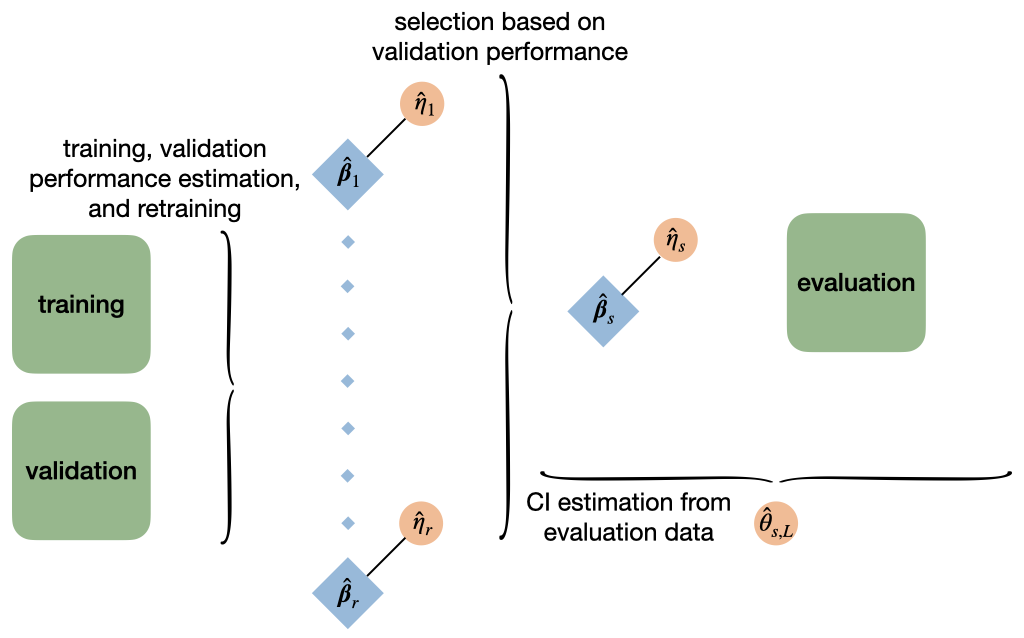}
\caption{Default evaluation pipeline, as predominantly recommended in the literature. Only a single model $\hat{\bm \beta}_s$ is selected for evaluation based on its validation performance $\hat{\eta}_s$. $\hat{\theta}_{s, L}$ is the lower confidence bound for that model's evaluation performance}
\label{fig:default-pipeline}
\end{center}
\end{figure}

We are particularly interested in the conditional prediction performance, that is the generalization performance of the model trained on the present sample. For that, in a model selection and performance estimation regime, the prevailing recommendation in the literature is to split the sample at hand into three parts, a training, validation, and evaluation set \citep{Goodfellow-2016, Hastie-2009, Japkowicz-2011, Murphy-2012, Raschka-2018}, see Figure \ref{fig:default-pipeline}. Depending on the specific selection rule, the training and validation set can sometimes be combined to form a learning set. For instance, this is true when cross-validation is used to identify promising models from a competition of models, based on their cross-validated prediction performance. Cross-validation using the entire sample at hand however is not a solution to our problem since it actually estimates the unconditional prediction performance, that is the average prediction performance of a model fit on other training data from the same distribution as the original data \citep{Bates-2021a, Hastie-2009}. There are examples in the literature how to correct for this unwanted behavior and report an estimate of the conditional performance \citep{Bates-2021a, Tsamardinos-2018}. Yet, we choose an approach that directly and inherently estimates the conditional performance. 

\subsection{Bootstrap Tilting Confidence Intervals}

Our proposed confidence bounds are obtained using bootstrap resampling. In particular, we use bootstrap tilting (BT), introduced by \cite{Efron-1981}, which is a general approach to estimate confidence intervals for some statistic $\theta = \theta (F)$ using an i.~i.~d.~sample $(y_1, y_2, \ldots, y_n)$ from an unknown distribution $F$. This statistic $\theta$ will later be our performance estimate of choice. Unlike many other bootstrap confidence intervals, BT estimates the distribution of $\hat{\theta} - \theta_0$ for some test value $\theta_0$. The lower confidence bound is then formed consisting of those values of $\theta_0$ that could not be rejected in a test of the null hypothesis $H_0 \colon \theta \leq \theta_0$. This way the distribution to resample from is consistent with the null distribution. In particular, this is achieved by reducing the problem to a one-parameteric family $(F_\tau)_\tau$ of distributions, where $\tau$ is called the \textit{tilting parameter} and $F_\tau$ has support on the observed data $\{ y_1, y_2, \ldots, y_n \}$. A specific value of $\tau$ induces nonnegative sampling weights $\bm p_\tau = (p_1(\tau), p_2(\tau), \ldots, p_n(\tau))$ such that $\sum_{i=1}^n p_i(\tau) = 1$. The tilting parameter $\tau$ is monotonically related to $\theta$ such that a specific value of $\tau$ corresponds to a specific value of $\theta_0$. 

In order to find a lower confidence bound $\theta_L$, we find the largest value of $\tau < 0$ such that the corresponding level $\alpha$ test still rejects $H_0$; this means $\theta_L$ is the largest value of $\theta_0$ such that, if the sample came from a distribution with parameter $\theta_0$, the probability of observing $\hat{\theta}$ or an ever larger value is $\alpha$, 
\begin{equation} \label{eq:tau-calib-generic}
P_{F_\tau} ( \theta \geq \hat{\theta} ) = \alpha.
\end{equation}

\begin{figure}[ht]
\begin{center}
\includegraphics[height=0.44\textheight]{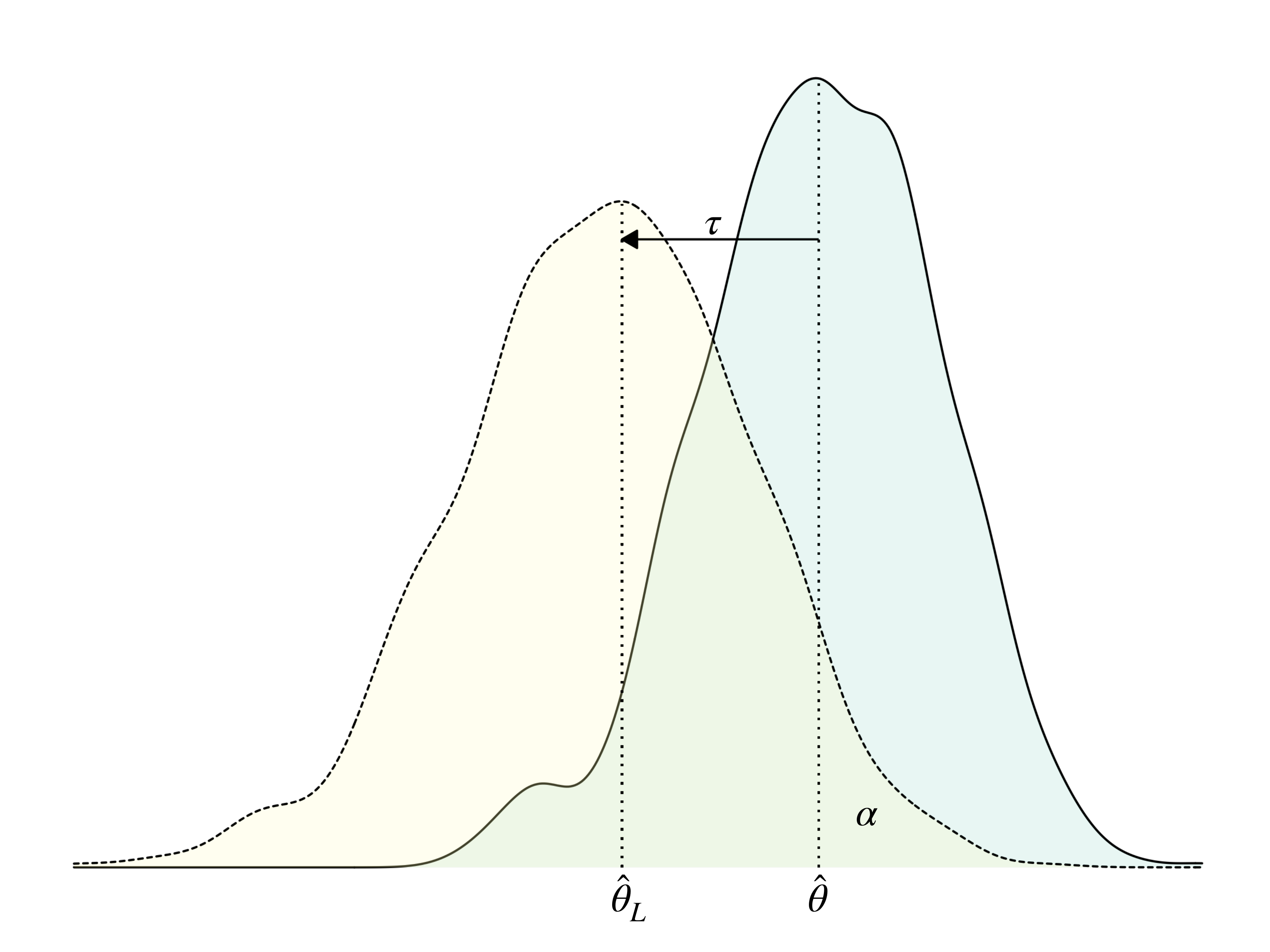}
\caption{BT confidence bound estimation. The solid-line distribution on the right represents $\hat{F}$, while the dashed-lined distribution on the left represents $\hat{F}_\tau$. BT finds a value for $\tau$ such that the probability under $\hat{F}_\tau$ to observe at least $\hat{\theta}$ is $\alpha$; this means the mass of the dashed-lined distribution that is to the right of $\hat{\theta}$ is equal to $\alpha$. The associated value $\hat{\theta}_L$ of $\theta$ under $\hat{F}_\tau$ is the desired lower confidence bound}
\label{fig:bt}
\end{center}
\end{figure}

Conceptually, for any given value of $\tau$, we need to sample from $F_\tau$ and check whether equation \eqref{eq:tau-calib-generic} holds true. This is both expensive and exposed to the randomness of repeated sampling. What we actually do is to employ an importance sampling reweighting approach as proposed by \cite{Efron-1981}. This allows us to find the lower bound $\theta_L$ using only bootstrap resamples from the observed empirical distribution $\hat{F}$. We reweight each resample $b = 1, \ldots, B$ with the relative likelihood $W_b(\tau) = \prod_{i=1}^n p_i(\tau) / \prod_{i=1}^n n^{-1}$ of the resample under $p_\tau$-weighted sampling relative to ordinary sampling with equal weights $n^{-1}$, and calibrate the tilting parameter $\tau < 0$ such that the estimated probability of observing at least $\hat{\theta}$ under the tilted distribution $F_\tau$ is $\alpha$, 
\begin{equation*}
\alpha = P_{F_\tau} (\theta(\hat{F}^*) \geq \hat{\theta}) = B^{-1} \sum_{b = 1}^B W_b(\tau) \, I \{ \hat{\theta}_b^* \geq \hat{\theta} \}, 
\end{equation*}
where $\hat{F}^*$ is the resampling empirical distribution. Then the value of the statistic $\theta$ that corresponds to that calibrated value of $\tau$ and the respective sampling weights $p_\tau$ is the desired lower confidence bound $\theta_L = \theta(p_\tau)$. Figure \ref{fig:bt} illustrates this idea.

The tilting approach does not work if the data $y_1 = \ldots = y_n$ to resample from is constant because then 
\begin{equation} \label{eq:bt-problem}
p_1(\tau) = \ldots = p_n(\tau) \quad \text{for any } \tau,
\end{equation}
and the empirical distribution cannot be tilted. This can for instance be an issue in binary classification, when the model perfectly predicts the true class labels. One option to deal with this issue is to switch to another (conservative) interval estimation method. In the aforementioned example this could for instance be a Clopper-Pearson lower confidence bound.

\begin{figure}[ht]
\begin{center}
\includegraphics[width=\textwidth]{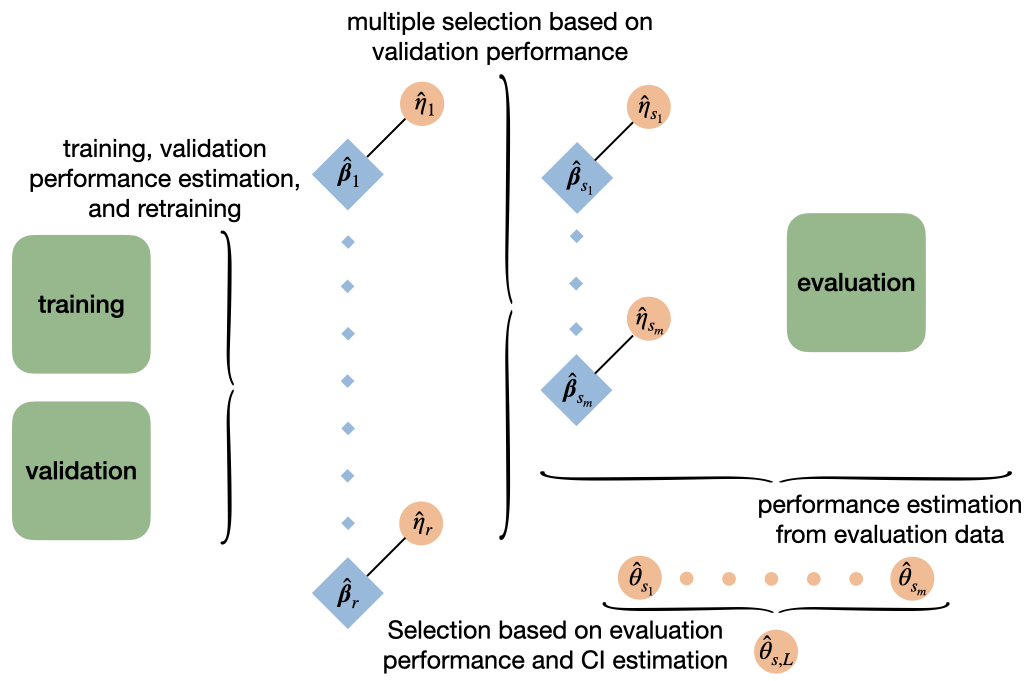}
\caption{Proposed evaluation pipeline. Multiple models $s_1, \ldots, s_m$ are selected for evaluation based on their validation performances $\hat{\eta}_{s_1}, \ldots, \hat{\eta}_{s_m}$, and a final model $s$ is selected based on the evaluation performances $\hat{\theta}_{s_1}, \ldots, \hat{\theta}_{s_m}$. This introduces a multiplicity problem concerning $\hat{\theta}_{s_1}, \ldots, \hat{\theta}_{s_m}$ and an upward bias in $\hat{\theta}_s$. We propose a lower confidence bound $\hat{\theta}_{s, L}$ that compensates for this multiplicity and bias}
\label{fig:proposed-pipeline}
\end{center}
\end{figure}

BT is known to be second-order correct and to work well for a single model, when no model selection is involved \citep{DiCiccio-1990, Hesterberg-1999}. However, in our proposed pipeline, multiple models are being evaluated, see Figure \ref{fig:proposed-pipeline}. Thus, we modify the BT routine and incorporate a maxT-type multiplicity control, which is a well-known standard approach in simultaneous inference \citep{Dickhaus-2014}. To the best of our knowledge, this is the first time that BT is extended to simultaneous inference and applied in a machine learning evaluation setup. Our approach enables us to simultaneously evaluate the conditional performances of multiple models and provide valid confidence bounds for them and in particular one for the final selected model. 

In the following, we consider a binary classification problem where a potentially large number $r$ of candidate models have already been trained and a number of promising models $s_1, \ldots, s_m$ have already been selected for evaluation, based on their validation performances $\hat{\eta}_{s_1}, \ldots \hat{\eta}_{s_m}$ and following some selection rule. We call this multitude of models selected for evaluation to be the set of \textit{preselected} models. In addition, we suppose that retraining of the preselected models on the entire learning data has already been performed, yielding models $\hat{\bm \beta}_{s_1}, \ldots, \hat{\bm \beta}_{s_m}$. Also, suppose that the associated performance estimates $\hat{\theta}_{s_1}, \ldots, \hat{\theta}_{s_m}$ have been obtained based on the predictions from the hold-out evaluation set, and a final model $s \in \{s_1, \ldots, s_m\}$ has been selected due to its evaluation performance $\hat{\theta}_s$ following some (other) selection rule. 

Section \ref{ch:method} has all the details to our proposed method. In Section \ref{ch:sim} we show a selection of results from our simulation experiments and a complete presentation can be found in the Supplementary Information. We apply our proposed approach on a real data set in Section \ref{ch:data-example}. Our presentation ends with a discussion in Section \ref{ch:summary}. 

\section{Method} \label{ch:method}


For brevity, let $j = 1, \ldots, m$ denote the preselected models instead of $s_1, \ldots, s_m$ and let $s \in \{1, \ldots, m\}$ denote the final selected model, that is the model with the most promising evaluation performance $\hat{\theta}_s$, which is a function of that model's evaluation predictions $\hat{y}_{1s}, \hat{y}_{2s}, \ldots, \hat{y}_{ns}$, where $n$ is the size of the evaluation set at hand. Note that this estimate $\hat{\theta}_s$ of generalizing prediction performance is subject to selection bias and therefore overly optimistic. To compute our proposed multiplicity-adjusted bootstrap tilting (MABT) lower confidence bound we only need these predictions $\hat{y}_{ij}$ from all of the competing preselected models $j = 1, \ldots, m$ in the evaluation set and the associated true class labels $y_i$, $i = 1, \ldots, n$. For instance, in case the performance measure of interest is prediction accuracy, the predictions are the predicted class labels; in case of the area under the receiver operating characteristic curve (AUC), they may be class probabilities. 

\subsection{Bootstrap Resampling}
Our proposed confidence bounds come from BT with a multiplicity adjustment due to the simultaneous evaluation of multiple candidate models. The general idea is to first estimate the tilted distribution under hypothetical values for the performance measure, and then to adjust for multiplicity in order to finally calibrate the tilting parameter $\tau$ accordingly. Note that no additional model training is needed here, once the models are preselected and retrained, using only the learning set. Thus, the predictions from the evaluation set can be understood as conditional on the trained models and on the learning data. We draw bootstrap resamples from the set of observation indices $\{1, \ldots, n\}$ in the evaluation set. For each of the candidate models and each of the resamples, we estimate the performance from this resample. This way, for each resample $b = 1, \ldots, B$ and each of the competing models $j = 1, \ldots, m$, we obtain a bootstrap performance estimate $\hat{\theta}_{bj}^*$. (Quantities that carry the $*$ subscript always indicate bootstrap quantities.) In the following, we use these bootstrap performance estimates to estimate various empirical cumulative distribution functions.

\subsection{Tilting}
The first of these empirical cumulative distribution functions stems from the estimation of conventional BT confidence intervals. From the bootstrap performance estimates $\hat{\theta}_{bs}^*$ of the final selected model $s$, we estimate the empirical \textit{tilted} cumulative distribution function 
\begin{equation} \label{eq:tilt-ecdf-selected}
\hat{F}_{s, \tau}^* (x) = \frac{1}{B} \sum_{b = 1}^B W_b (\tau) \, I \{ \hat{\theta}_{bs}^* \leq x \}, 
\end{equation}
where $I$ is the indicator function, with importance sampling weights $W_b(\tau)$, as introduced in Section \ref{sec:introduction}. We plug the observed performance $\hat{\theta}_s$ of the final selected model $s$ into $\hat{F}_{s, \tau}^*$ to get $\hat{F}_{s, \tau}^* (\hat{\theta}_s)$. Later, we plug this value into another distribution function that we derive next. This will provide the multiplicity correction that we need to account for the evaluation of multiple models.

\subsection{Multiplicity Correction and Calibration}
Since the performance estimates $\hat{\theta}_{bj}^*$ all come from different and not necessarily comparable models, we need to transform them appropriately to bring them to a comparable scale and compute a multiplicity-adjusted ($1$ minus) p-value that can be used later on to calibrate the tilting parameter $\tau$ and finally obtain the desired lower confidence bound. For each of the models $j$, we estimate the empirical cumulative distribution function  
\begin{equation*}
\hat{F}_j^* (x) = \frac{1}{B} \sum_{b = 1}^B I \{ \hat{\theta}_{bj}^* \leq x \} 
\end{equation*}
from the bootstrap performance estimates $\hat{\theta}_{bj}^*$. Plugging these into $\hat{F}_j^*$ for each $j$ yields transformed bootstrap performance estimates $\hat{u}_{bj}^*$ that are now uniformly distributed on the unit interval. Similar to a maxT multiplicity correction, for each bootstrap resample $b$, among the transformed estimates $\hat{u}_{bj}^*$, we now identify the maximum estimate $\hat{u}_{b, \max}^* = \max_{j=1}^m \hat{u}_{bj}^*$. From these maximum values, we estimate
\begin{equation} \label{eq:max-ecdf}
\hat{F}_{max}^* (x) = \frac{1}{B} \sum_{b = 1}^B I \{ \hat{u}_{b, \max}^* \leq x \}
\end{equation}
and refer to it as the maximum empirical cumulative distribution function.

\subsection{Lower Confidence Bound}
In the previous paragraphs, we estimated two empirical cumulative distribution functions, $\hat{F}_{s, \tau}^*$ and $\hat{F}_{max}^*$ in equations \eqref{eq:tilt-ecdf-selected} and \eqref{eq:max-ecdf}, respectively. The former estimates the tilted cumulative distribution function while the latter corrects for multiplicity. We next bring those two together and state the calibration task to obtain a conditional lower confidence bound for the performance of the final selected model $s$ from our proposed MABT approach: Find a value of the tilting parameter $\tau < 0$ such that
\begin{equation} \label{eq:mabt-calibration-task}
\hat{F}_{max}^* [ \hat{F}_{s, \tau}^* (\hat{\theta}_s) ] = 1 - \alpha.
\end{equation}
Once this specific value for $\tau$ has been found, we finally obtain the desired lower confidence bound via 
\begin{equation*}
\hat{\theta}_{s, L} = \theta ( \hat{F}_{s, \tau}^* ).
\end{equation*}

We implemented this numerically as a root-finding problem which calibrates the tilting parameter $\tau$ conservatively, that means equation \eqref{eq:mabt-calibration-task} holds with $\geq$. An R implementation of our proposed confidence bounds can be accessed via a public GitHub repository at \url{https://github.com/pascalrink/mabt}. It is able to compute our proposed confidence bounds for the prediction accuracy and the AUC of a prediction model. 

\begin{figure}[ht]
\begin{center}
\includegraphics[width=\textwidth]{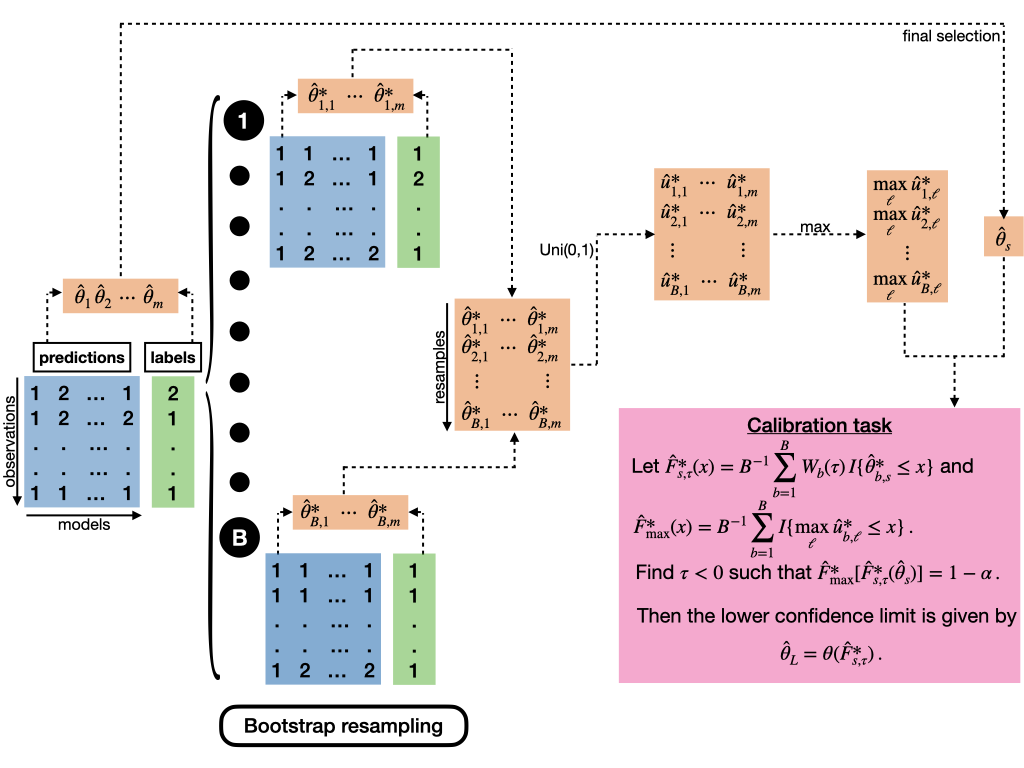}
\caption{Schematic illustration of the proposed MABT approach to estimate confidence bounds for prediction accuracy}
\label{fig:proposed-method}
\end{center}
\end{figure}

Figure \ref{fig:proposed-method} illustrates our proposed approach in case the performance measure is prediction accuracy. In summary, we estimate a BT confidence bound using an adjusted significance level due to the maximum distribution $\hat{F}_{\max}^*$ of all the preselected prediction models $j = 1, \ldots, m$. This yields simultaneous confidence bounds $(\hat{\theta}_{1, L}, \ldots \hat{\theta}_{m, L})$ for all the preselected models $j$, that means equation \eqref{eq:fwer} holds. This way, due to equation \eqref{eq:from-fwer-follows-marginal}, we obtain a valid confidence bound for the prediction performance of any model $j = 1, \ldots, m$ and hence a valid confidence bound $\hat{\theta}_{s, L}$ for the selected model $s$. 

\section{Simulation Experiments} \label{ch:sim}

We now investigate the goodness of our proposed confidence bounds in extensive simulation experiments and compare them to existing standard approaches in terms of four aspects: coverage probability, size of lower confidence bound, true performance of the final selected model, and the distance between the true performance and the lower bound, which we call \textit{tightness} for brevity. We expect our procedure to produce models with better true predictive performance than standard methods due to the gainful way we select models for evaluation (see Figure \ref{fig:proposed-pipeline}) in comparison to the default pipeline (see Figure \ref{fig:default-pipeline}). On the other hand, due to the present multiplicity that we need to account for, the question is if this reflects in larger lower confidence bounds, as this is the main quantity when reporting. In other words, there is no real advantage to have found a better prediction model if we are unable to identify it as such. 

The R code used for our simulation experiments and for all the associated figures can be found in a publicly accessible GitHub repository at \url{https://github.com/pascalrink/mabt-experiments}.

\subsection{Setup}

\subsubsection{Data Generation} \label{sec:data-gen}

Using a standard 50\%/25\%/25\% split for the samples sizes of 200 and 400 observations for prediction accuracy and, because it is a more complex measure, 400 and 600 observations for the AUC, respectively, we sample the training, validation, and evaluation data as well as another larger data set of 20,000 observations all from the same distribution. The latter serves as a ground truth from which we derive the true model performances. We consider two different setups A and B how to generate the features and how to get the true class labels from them. They represent different data complexities in some way. Nonetheless, both setups lead to balanced classes. 

In the first, simpler, and somewhat synthetic case A,  we draw uncorrelated random numbers from the standard normal distribution and put them into the feature matrix $\bm X = (\bm x_i)_i$ with columns $\bm x_i$. We specify a sparse true coefficient vector $\bm \beta$ with only 1\% non-zero coefficients and obtain the true class labels using the inverse logit function and a vector of uncorrelated observations $u_i$ that are uniform on the unit interval, 
\begin{equation*}
y_i = I \Bigl\{ \frac{1}{1 + \exp(- \bm x_i \bm \beta)} \geq u_i \Bigr\}.
\end{equation*}
Specifically, we draw 1000 features and choose $\bm \beta = c \cdot (1, 1, \ldots, 1, 0, 0, \ldots, 0)$ to have only $10$ nonzero entries, with signal strength $c = 2$. 

In case B, we use the \texttt{twoClassSim} R function from the \texttt{caret} R package \citep{Kuhn-2008}. As to the true class labels, this function generates a feature matrix that includes linear effects, non-linear effects, and noise variables, each both uncorrelated and correlated with constant correlation $\rho = 0.8$, and $1\%$ mislabeled data. We try to mimic a more complex case here, which is closer to real-world applications than case A.

\subsubsection{Model Training}

As to model training, we train lasso models with 100 equidistant values for the tuning parameter $\lambda$ between zero and the maximum regularization value $\lambda_{\max} = \min \{ \lambda > 0 \colon \hat{\bm \beta}_\lambda = \bm 0 \} $, which is the smallest tuning parameter value such that none of the features is selected into the model, where $\hat{\bm \beta}_\lambda$ is the estimate of the true vector $\bm \beta$ of coefficients from a lasso regression with regularization parameter value $\lambda$. Note that $\lambda_{\max}$ depends on the input training data and we leave its computation to the \texttt{glmnet} R function from the \texttt{glmnet} R package \citep{Friedman-2010}. 

\subsubsection{Performance Estimation} 

In our proposed pipeline (see Figure \ref{fig:proposed-pipeline}), performance estimation happens at two different stages. First, the performances $\hat{\eta}_1, \ldots, \hat{\eta}_r$ of all of the candidate models is estimated. One way to do this is to train the models using the training data and estimate their performances using the validation data. Another option is to perform ten-fold cross-validation on the entire learning data, which is the combination of training and validation data. We expect these estimates to be less dependent on the split of the data into training and validation sets and, thus, to lead to better final selections. For brevity, we only compute the cross-validated estimates in case of the AUC. Note that when using cross-validation the resulting performance estimates are neither conditional on the trained model nor validation performances in the sense of the non-cross-validated estimates; they are averaged unconditional performances over the ten folds. However, we use these performances only to identify promising models to preselect for evaluation later on.

Once models are preselected for evaluation based on their (cross-) validation performances, they are refitted using the entire learning data before their generalization performances are estimated using the evaluation data. This is the second and final round of performance estimation. 

\subsubsection{Model Selection}

As with performance estimation, model selection happens at two different stages, too, see Figure \ref{fig:proposed-pipeline}. First, promising models are identified based on their validation performance and preselected for evaluation. Second, a final model $s$ is selected among them, based on its evaluation performance $\hat{\theta}_s$. The two selection rules do not necessarily need to be the same. In our simulation experiments, based on their validation performance, we preselect for evaluation either the single best model, or the top 10\% of the models. In case the cross-validation performance is used for selection, in addition to the \textit{single best} and the \textit{top 10\%} selection rule, we select all the models with cross-validation performance within one cross-validation standard error of the best model. The preselected models are then refitted using all of the training and validation data. In any case, in the final selection stage, we select the one model $s$ with the best evaluation performance $\hat{\theta}_s = \max_{i = 1}^m \hat{\theta}_i$ and report a lower confidence bound $\hat{\theta}_{s, L}$ for it. 

\subsubsection{Confidence Bounds}

We compute our MABT lower confidence bounds using 10,000 bootstrap resamples when the performance measure of interest is the prediction accuracy, and 2000 resamples in case of AUC, for computational brevity. We use the R function \texttt{auc} from the R package \text{pROC} to estimate the AUC \citep{Robin-2011}. Table \ref{table:sim-configs} provides a summary of the confidence bounds estimated for each configuration. In case of prediction accuracy, we compare our proposed bounds to a number of existing standard approaches: the Wald normal approximation interval, which is known to sometimes struggle to reach the nominal significance level, especially when sample size is small; the Wilson interval, which is an improvement over the Wald interval in many respects as it allows for asymmetric intervals, incorporates a continuity correction, and can also be used when sample size is small; the Clopper-Pearson (CP) exact interval, which uses the binomial and, thus, the correct distribution rather than an approximation to it; and the default BT confidence interval as presented in Section \ref{sec:introduction}. In case of the AUC, we again take the BT confidence interval for comparison, as well as DeLong intervals, which are the default choice for an asymptotic interval here, and Hanley-McNeil (HM) intervals, which use a simpler a variance estimator. To compute the DeLong intervals we use the R function \texttt{ci.auc} from the \texttt{pROC} R package \citep{Robin-2011}. 

In those scenarios where multiple models are evaluated simultaneously, a multiplicity adjustment is necessary. Our proposed confidence bounds control for such multiplicity, whatever the selection rule applied. For the standard methods in the competition however we apply the \Sidak -correction and estimate the confidence bound using a reduced and adjusted significance level of $\alpha_{\text{\Sidak}} (m) = 1 - (1 - \alpha)^{1 / m}$, where $m$ is the number of preselected models. We can safely assume that the predictions from the various candidate models are not negatively dependent and, thus, we choose the \Sidak ~over the Bonferroni adjustment \citep{Dickhaus-2014}, since they are less conservative. 

\begin{table}[ht]
\centering
\begin{tabular}{lll}
\toprule
Selection rule        & Prediction accuracy          & AUC                    \\ \midrule
Single best           & BT       & BT                     \\
                      & Wald                         & DeLong                 \\
                      & Wilson                       & HM          \\
                      & CP         &                        \\
Top 10\%              & MABT                         & MABT                   \\
                      & BT + \Sidak                  & BT + \Sidak            \\
                      & Wald + \Sidak                & DeLong + \Sidak        \\
                      & Wilson + \Sidak              & HM + \Sidak \\
                      & CP + \Sidak                  &                        \\ 
Within 1 SE           & MABT  & MABT                   \\
                      & BT + \Sidak                  & BT + \Sidak            \\
                      & Wald + \Sidak                & DeLong + \Sidak        \\
                      & Wilson + \Sidak              & HM + \Sidak \\
                      & CP + \Sidak                  &                        \\ \bottomrule
\end{tabular}
\vspace{11pt}
\caption{Listing of the considered interval procedures per performance measure and selection rule. The \textit{within 1 SE} selection rule is only considered when cross-validation is used to estimate the validation performance}
\label{table:sim-configs}
\end{table}

\subsection{Results}

\begin{figure}[ht]
\begin{center}
\includegraphics[height=0.44\textheight]{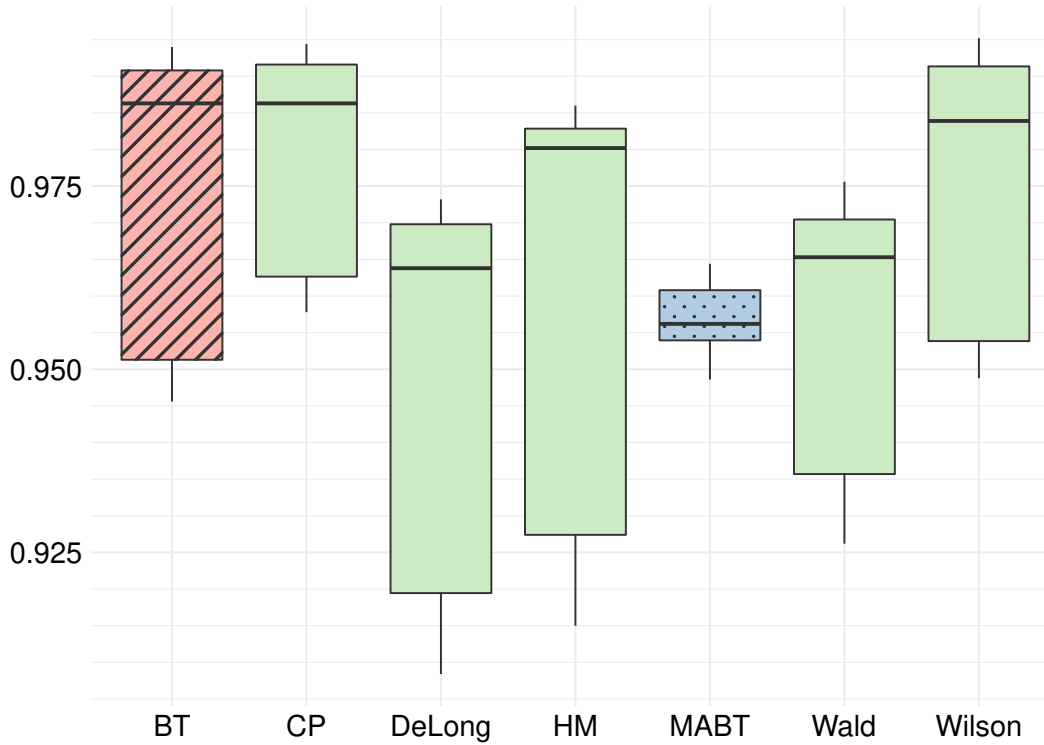}
\caption{Observed coverage probability over all the simulation experiments. The proposed MABT bounds are the most reliable ones regarding coverage probability; in the majority of cases they surpass the desired coverage level and the coverage probability varies the least}
\label{fig:comp-all-coverage}
\end{center}
\end{figure}

Next we present the results from 5000 simulation runs for each combination of simulation parameters. Figure \ref{fig:comp-all-coverage} presents the overall observed coverage probabilities of the seven interval methods over all simulation experiments, which means over all performance measures (prediction accuracy, AUC), selection rules (\textit{single best}, \textit{top 10\%}, \textit{within 1 SE}), sample sizes (200, 400 for prediction accuracy and 400, 600 for AUC, that means evaluation sample sizes 50, 100 and 100, 150, respectively), for both validation performance estimation variants (with and without cross-validation), and both feature generation methods (A, B). Overall, our proposed approach yields the most reliable observed coverage probabilities. (In none of our simulation runs we have encountered the issue described in Equation \ref{eq:bt-problem}.)

For brevity, we only present a selection of more detailed results here, which are the most competitive approaches in terms of coverage probability and size of the lower confidence bounds. The individual arguments why we dismiss certain results from presentation here can be found in the Supplementary Information. For prediction accuracy, this leaves us with the following competition: Wilson unadjusted confidence bounds for the single best performing model; BT confidence bounds for the single best performing model; our proposed MABT confidence bounds for the best model of the \textit{within 1 SE} selection of models. For AUC, the competition is the same as for prediction accuracy with DeLong replacing Wilson. In any case, for preselection, the validation performance is computed using the cross-validation estimate from the learning data. Since we split the data into 50\% training, 25\% validation, and 25\% evaluation data, for the selected results presented here, this is effectively a 75\%/25\% learning/evaluation split because for cross-validation we use both the training and validation data. 

\begin{figure}
\begin{center}
\includegraphics[height=0.44\textheight]{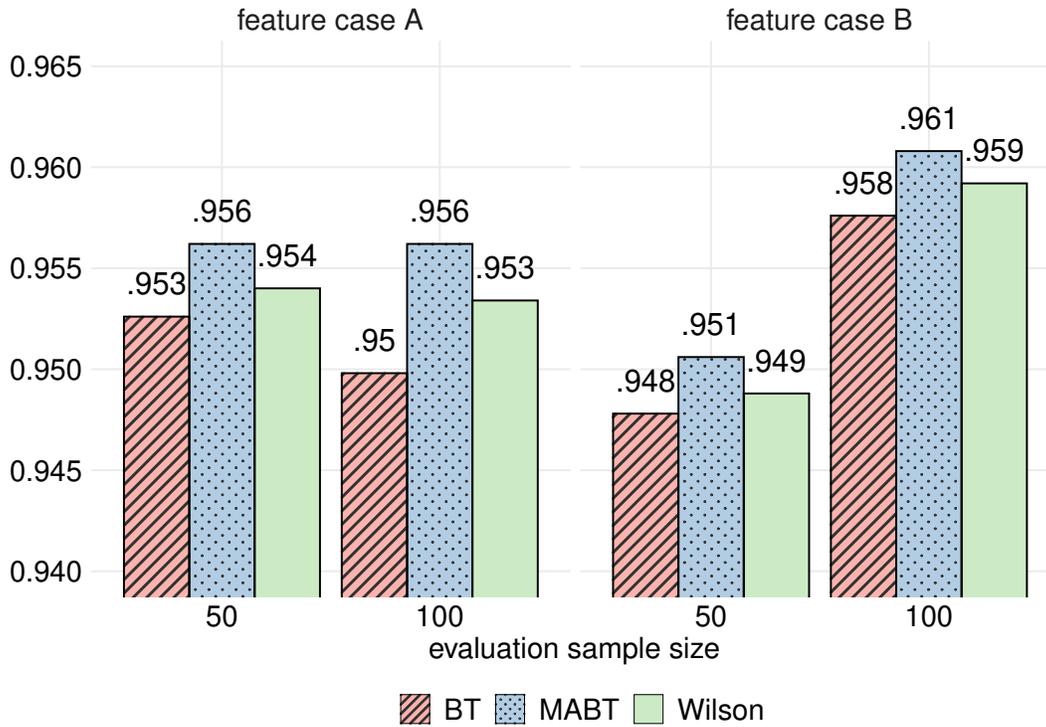}
\end{center}
\caption{Observed coverage for prediction accuracy. All the methods yield acceptable coverage probability, with MABT yielding slightly conservative bounds}
\label{fig:coverage-acc}
\end{figure}

\begin{figure}
\begin{center}
\includegraphics[height=0.44\textheight]{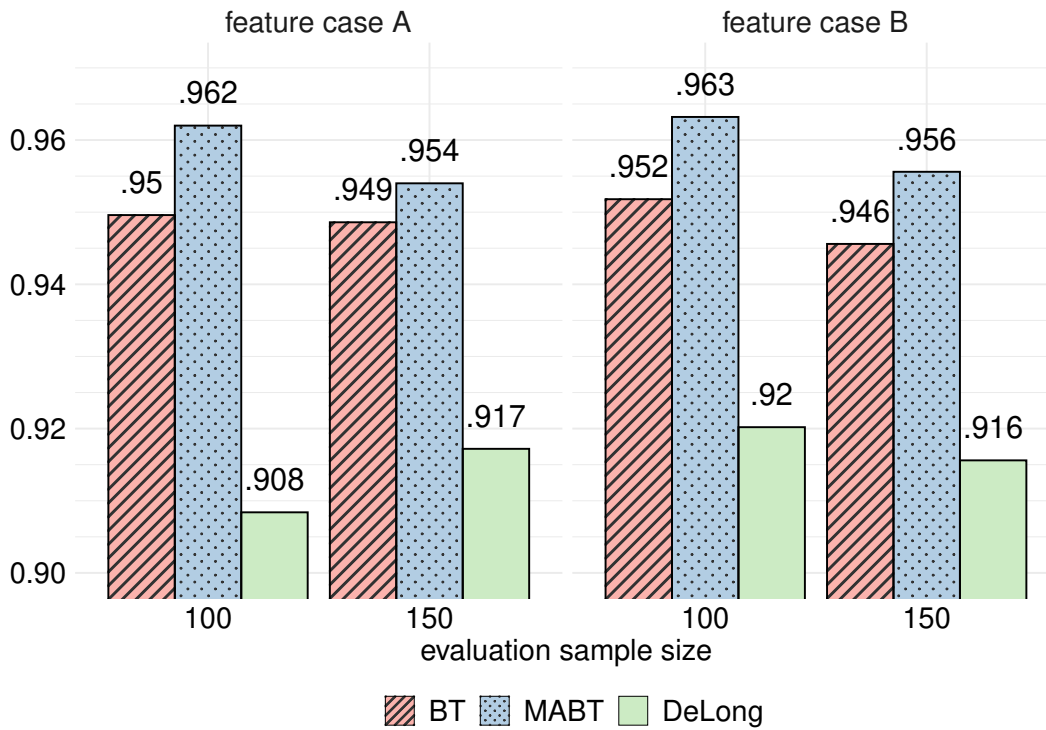}
\end{center}
\caption{Observed coverage for AUC. DeLong bounds are clearly too liberal; the coverage probability of the BT bounds is close to the desired level and the MABT bounds are again slightly conservative}
\label{fig:coverage-auc}
\end{figure}

\begin{figure}
\begin{center}
\includegraphics[height=0.44\textheight]{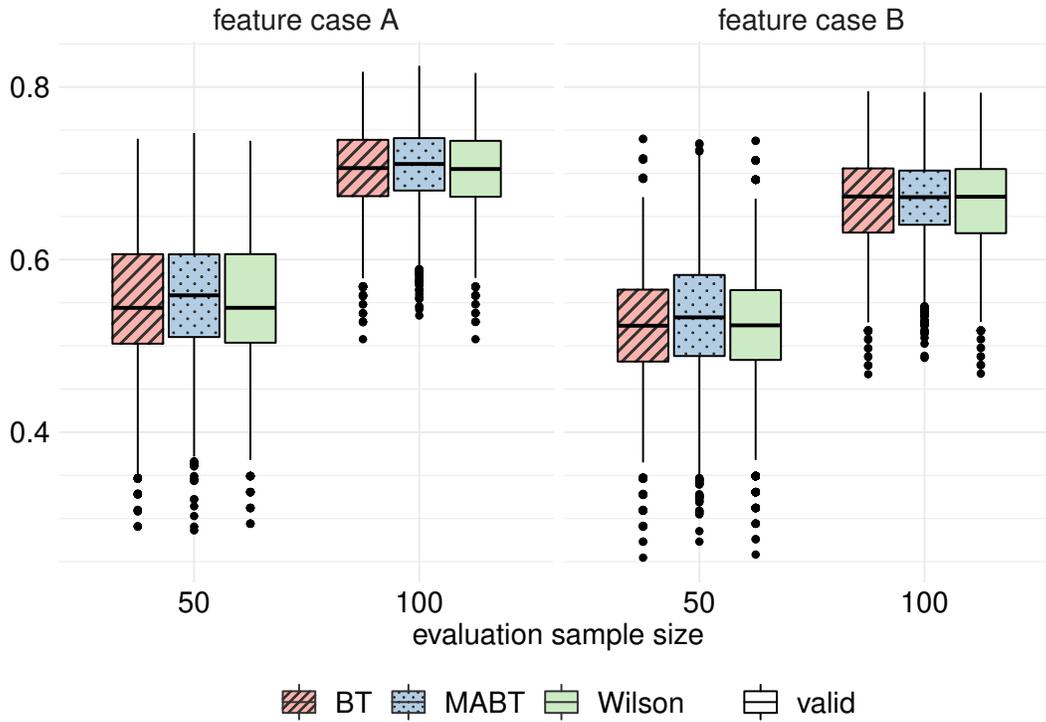}
\end{center}
\caption{Lower confidence bounds for prediction accuracy. MABT bounds are the largest in size}
\label{fig:lower-bounds-acc}
\end{figure}

\begin{figure}
\begin{center}
\includegraphics[height=0.44\textheight]{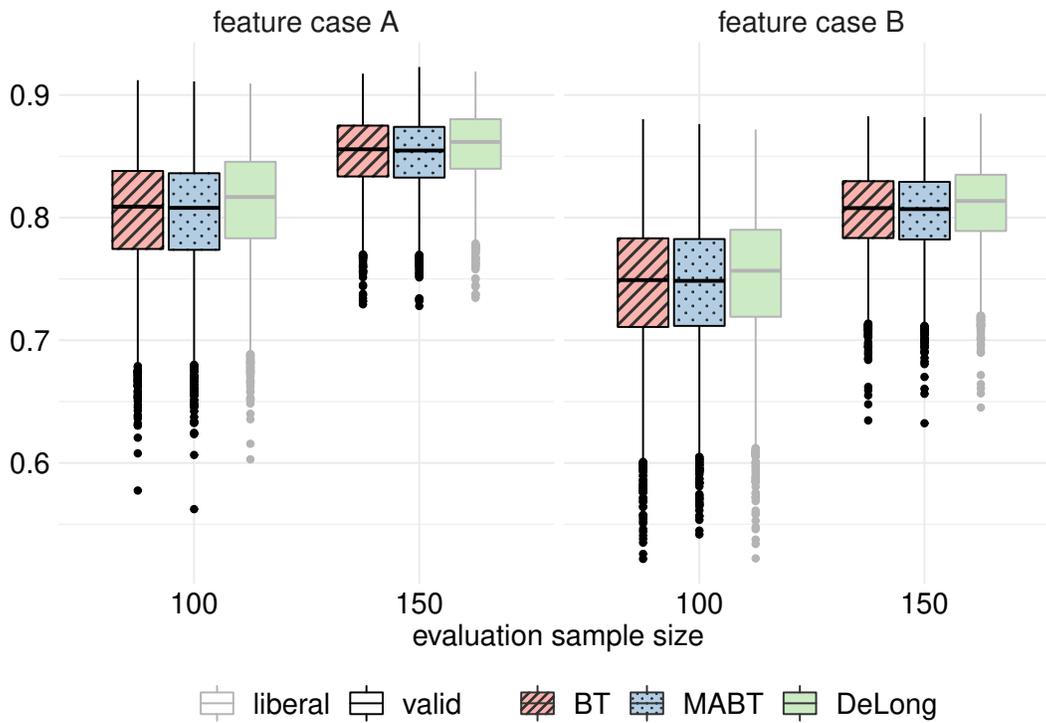}
\end{center}
\caption{Lower confidence bounds for AUC. BT and MABT bounds are similar regarding size, with the DeLong bounds being larger, but they are too liberal}
\label{fig:lower-bounds-auc}
\end{figure}

\begin{figure}
\begin{center}
\includegraphics[height=0.44\textheight]{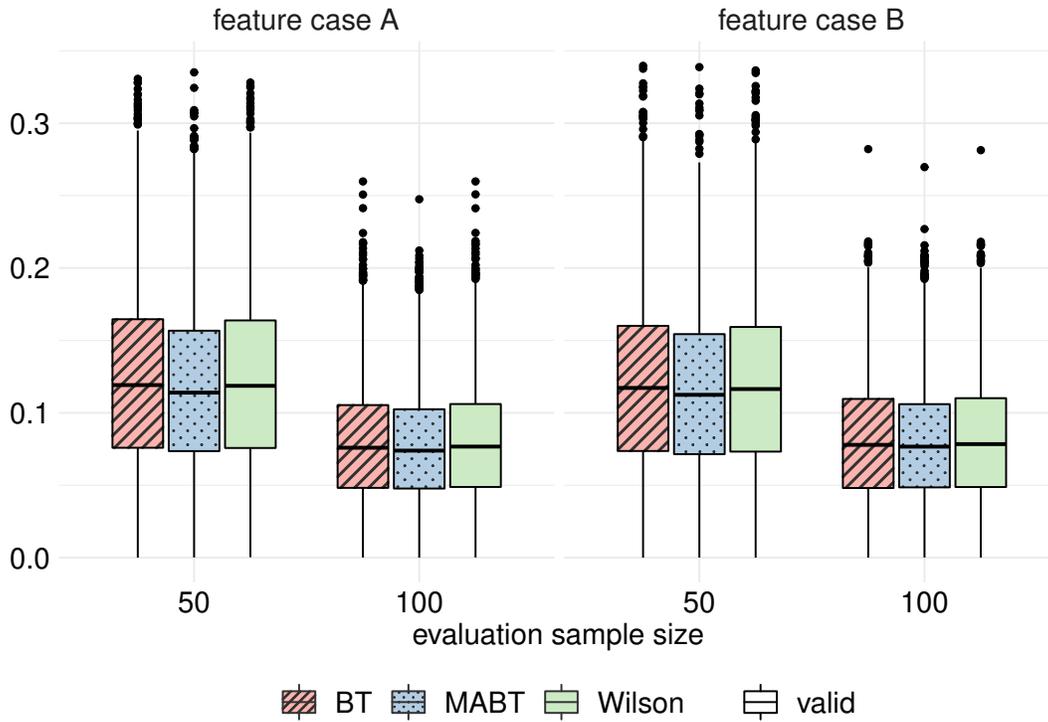}
\end{center}
\caption{Tightness for prediction accuracy. MABT are slightly tighter than the BT and Wilson bounds}
\label{fig:tightness-acc}
\end{figure}

\begin{figure}
\begin{center}
\includegraphics[height=0.44\textheight]{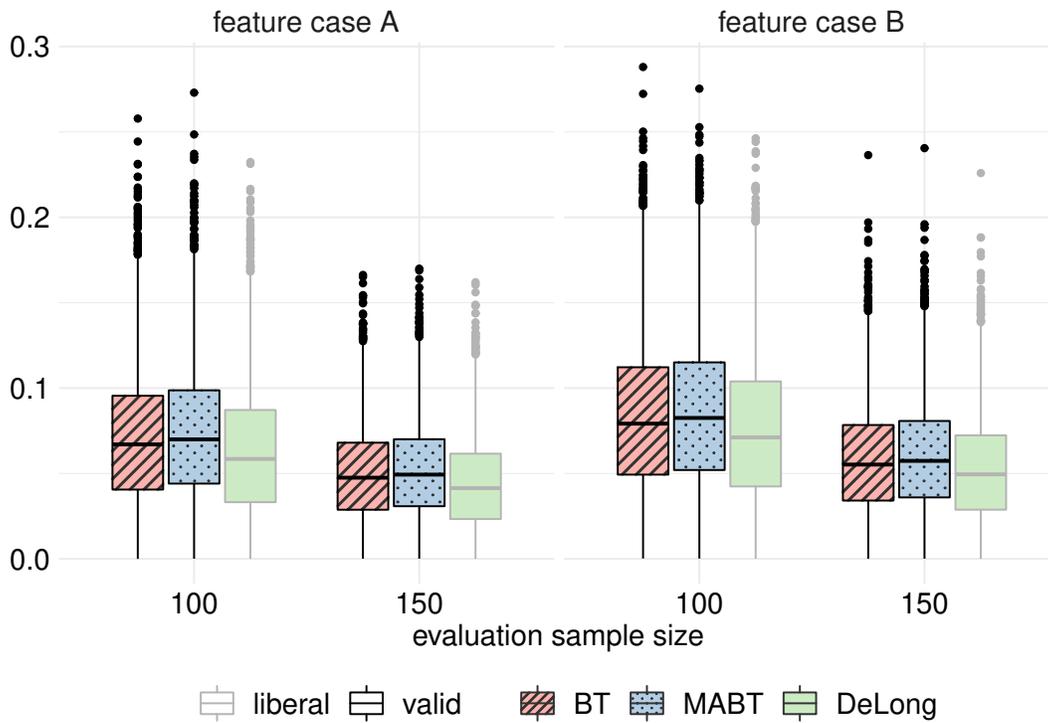}
\end{center}
\caption{Tightness for AUC. BT bounds are tighter than the MABT bounds, as are the DeLong bounds, but they are too liberal}
\label{fig:tightness-auc}
\end{figure}

\begin{figure} 
\begin{center}
\includegraphics[height=0.44\textheight]{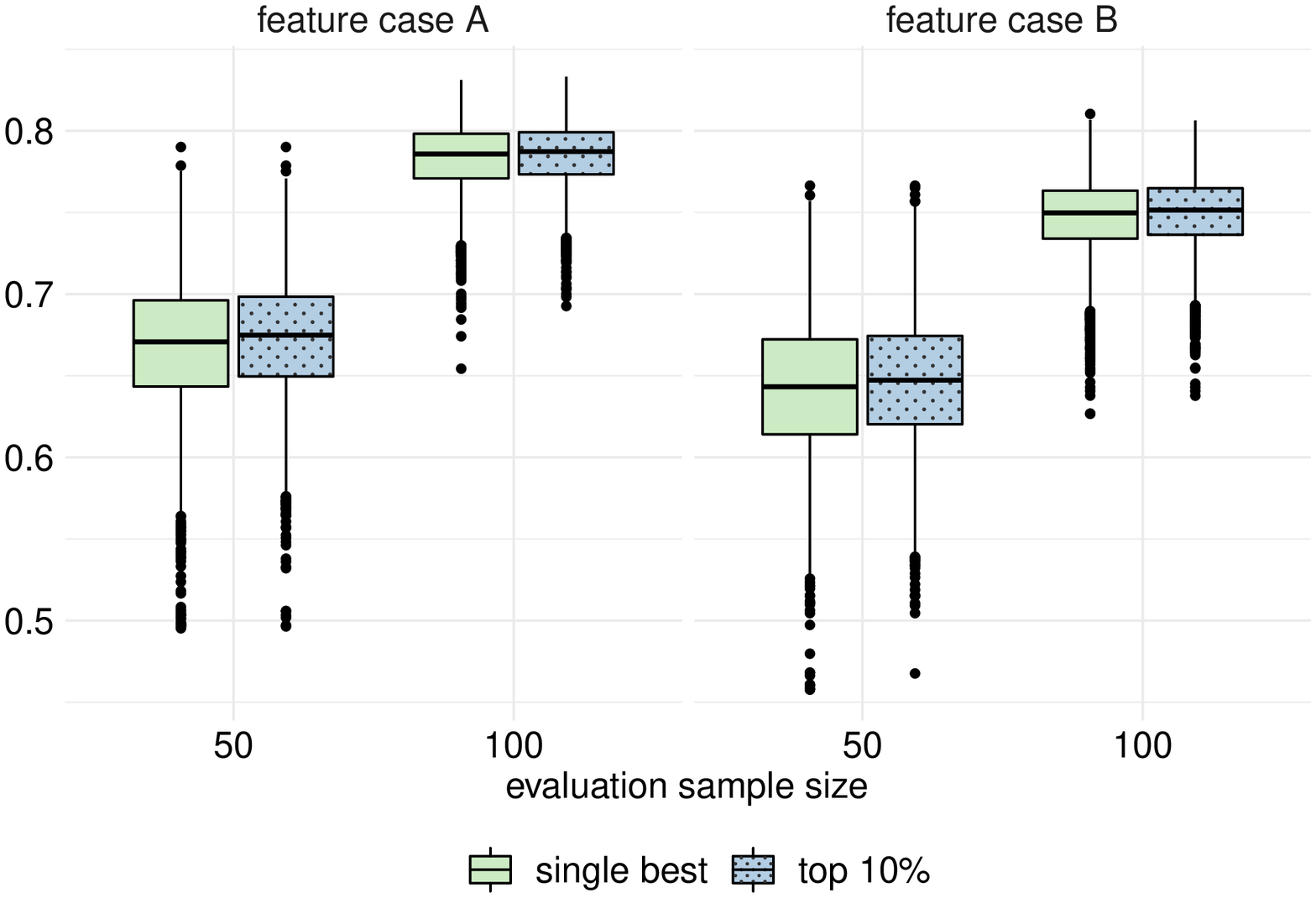}
\end{center}
\caption{True prediction accuracy of the final selected model. Evaluation of multiple models yields models with slightly better prediction performance compared to the default pipeline}
\label{fig:true-performance-acc}
\end{figure}

\begin{figure} 
\begin{center}
\includegraphics[height=0.44\textheight]{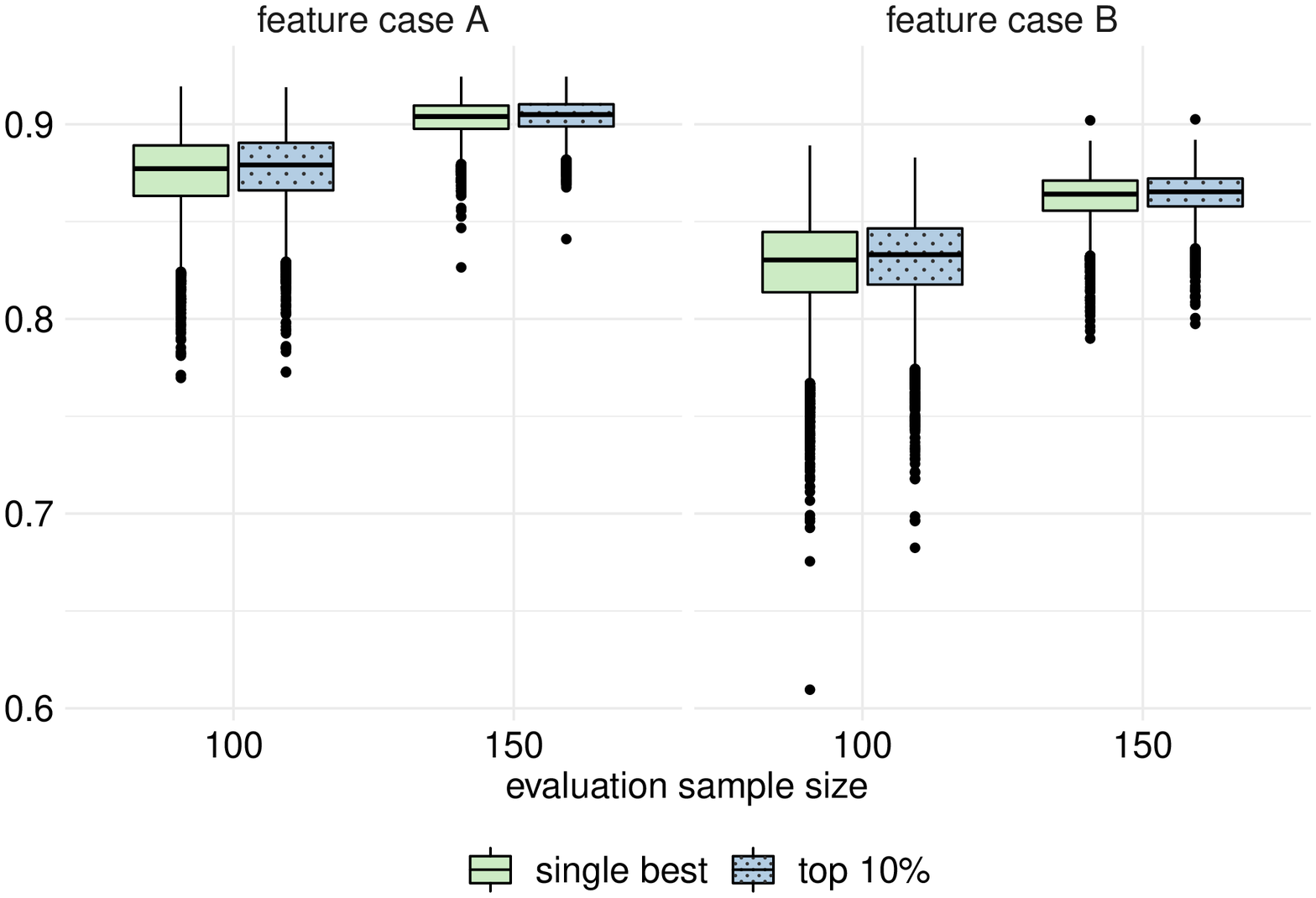}
\end{center}
\caption{True AUC of the final selected model. Evaluation of multiple models yields models with slightly better prediction performance compared to the default pipeline}
\label{fig:true-performance-auc}
\end{figure}

In the results we see that our proposed confidence bounds are somewhat conservative in the simpler setups, that is feature case A with prediction accuracy, and less conservative in the more complex setups, that is feature case B or in case of AUC, see Figures \ref{fig:coverage-acc} and \ref{fig:coverage-auc}. At the same time, the adverse effect of conservatism is not pronounced and does not directly translate into smaller lower bounds, see Figures \ref{fig:lower-bounds-acc} and \ref{fig:lower-bounds-auc}. Similar observations have already been made in \cite{Hall-1988}, where other conservative bootstrap intervals were not directly related to larger intervals. Also, the confidence bounds from our proposed method are less variable than the competing methods. Regarding tightness, we cannot draw a final conclusion, see Figures \ref{fig:tightness-acc} and \ref{fig:tightness-auc}. Note however that in some AUC simulation configurations the BT and the DeLong method yield too liberal intervals such that the comparison of our proposed confidence bounds to those in terms of size of the lower bound and tightness is unfair. (Too liberal intervals are identified as those whose coverage falls below $0.9469 = 1 - \alpha - \sqrt{(1-\alpha) \cdot \alpha / 5000}$, which is the desired coverage probability minus one standard error due to the finite number of simulation runs.) Nevertheless, especially in case of the smaller sample size, there is a visible gain in final model performance, see Figures \ref{fig:true-performance-acc} and \ref{fig:true-performance-auc}.





\section{Data Example} \label{ch:data-example}

In this Section, we apply the proposed approach on the Breast Cancer Wisconsin Diagnostic Data Set from the UCI Machine Learning Repository \citep{Dua-2017}, which consists of 698 observations on nine characteristics of cell nuclei present in digitized images of a fine needle aspirate of breast mass. The goal is to predict malignant tissue, which is the label in 241 cases. We split the data randomly into 523 observations for learning (75\%) and 175 for evaluation (25\%). In the learning phase, 100 lasso models with equidistant tuning parameters between $0$ and $\lambda_{\max} = 0.3829133$ are trained and ten-fold cross-validated performances on the learning data are obtained for each of the candidate models. We employ the \textit{single best}, \textit{top 10\%}, and \textit{within 1 SE} selection rules on the validation performances, refit the preselected models on the entire learning set, and select the one model with the highest evaluation performance among them for confidence bound estimation. We compute all the different interval procedures from all three selection strategies for comparison here (see Table \ref{table:sim-configs}). For the bootstrap-based confidence bounds we use 10,000 bootstrap resamples in case of prediction accuracy and 2000 resamples in case of AUC. In case of the prediction accuracy, for the \textit{within 1 SE} and \textit{top 10 \%} selection rules, the BT, CP, Wald, and Wilson confidence bounds use \Sidak -adjusted significance levels of about $0.0043$ ($m = 12$ preselected models) and $0.0085$ ($m = 6$); in case of the AUC, these levels are $0.0057$ ($m = 11$) and $0.0023$ ($m = 22$), respectively.

Table \ref{tab:data-example-accuracy-results} presents the results for the accuracy. The proposed MABT confidence bounds turn out to be among the largest; the MABT confidence bound from the \textit{within 1 SE} selection is only being outperformed in the default pipeline by the Wald confidence bound, which however are questionable in terms of coverage probability. In Table \ref{tab:data-example-auc-results} the results for the AUC are shown. The MABT bounds are about as large as the BT bound from the default pipeline and only slightly smaller than the DeLong and HM bounds. Note however that the latter two turn out to be too liberal in our simulations and, thus, the DeLong and HM bounds are questionable.  

\begin{table}[ht]
\centering
\begin{tabular}{@{}lccc@{}}
\toprule
                              & Single best & Top 10\%     & Within 1 SE  \\ \midrule \midrule
No.~$m$ of preselected models & 1           & 12           & 6            \\ \midrule \midrule
Performance                   &             &              &              \\ \midrule
Validation                    & 96.0       & 95.4---96.0    & 95.8---96.0 \\
Evaluation                    & 96.0       & 95.4---96.0    & 96.0---96.0 \\ \midrule \midrule
Lower confidence bound        &            &               &              \\ \midrule
BT                            & 92.9       & 90.7          & 91.3        \\
CP                            & 92.6       & 90.4          & 91.0        \\
MABT                          & ---        & 92.0          & 92.8        \\
Wald                          & 93.6       & 92.1          & 92.4        \\
Wilson                        & 92.8       & 90.1          & 90.8       \\ \bottomrule
\end{tabular}
\vspace{11pt}
\caption{Lower confidence bounds for the prediction accuracy. Our proposed MABT bound from the \textit{within 1 SE} selection rule is only inferior to the default pipeline Wald bound, which however is questionable in terms of coverage probability}
\label{tab:data-example-accuracy-results}
\end{table}

\begin{table}[ht] 
\centering
\begin{tabular}{@{}lccc@{}}
\toprule
                              & Single best & Top 10\%     & Within 1 SE  \\ \midrule \midrule
No.~$m$ of preselected models & 1           & 11           & 22           \\ \midrule \midrule
Performance                   &             &              &              \\ \midrule
Validation                    & 99.2        & 99.2---99.2   & 99.1---99.2 \\
Evaluation                    & 99.4        & 99.4---99.4   & 99.4---99.4 \\ \midrule \midrule
Lower confidence bound        &             &              &              \\ \midrule
BT                            & 98.4        & 97.8         & 97.7        \\
DeLong                        & 98.8        & 98.6         & 98.9        \\
HM                            & 98.7        & 98.8         & 98.8        \\
MABT                          & ---         & 98.3         & 98.4        \\ \bottomrule
\end{tabular}
\vspace{11pt}
\caption{Lower confidence bounds from each approach for the AUC. Our proposed MABT bound are slightly smaller than the DeLong and the HM bounds, which however are both questionable in terms of coverage probability}
\label{tab:data-example-auc-results}
\end{table}

In summary, while our proposed MABT confidence bounds show themselves reliable with regard to coverage probability in our simulation experiments (see Section \ref{ch:sim}), in this real-data example they turn out competitive regarding the size of the lower confidence bound.


\section{Discussion} \label{ch:summary}

We proposed MABT confidence bounds that allow for valid inference when more than a single candidate predictive model is evaluated. This is especially beneficial in situations where data is scarce as it eases the allocation of data towards learning and evaluation. Other methods in the literature either do not estimate conditional performance or do so only indirectly (for instance, cross-validation and nested cross-validation as in \cite{Bates-2021a}). This is different with the proposed approach, which directly estimates the conditional performance by resampling from the model's predictions from the evaluation set. 

The perhaps most prominent advantage of our proposed approach is that it can be universally applied. In principle, it works with any selection rule, any competition of prediction models, even from different model classes, and any performance measure as long as there is a version of it which accepts weights. Also, no additional model training is necessary, such that the additional computational burden is little. 

In our simulation experiments, our proposed confidence bounds turn out to be reliable regarding coverage probability while still offering comparably large lower confidence bounds as competing methods. Also, the evaluation of multiple candidate models instead of a single one can lead to better final model selections. But with increasing sample size, these advantages decrease. In a real data example, the proposed confidence bounds perform comparably well to the competition, and were only outperformed by methods that show themselves to be to be too liberal in some situations.

\section*{Acknowledgements and Funding}

The authors acknowledge the provision of the Breast Cancer Wisconsin Diagnostic Data Set at the UCI Machine Learning Repository (\cite{Dua-2017}). This project was funded by the Deutsche Forschungsgemeinschaft (DFG) project number 281474342.

\bibliography{main}

\end{document}